\documentclass[runningheads]{llncs}
\usepackage{graphicx}
\usepackage{subfigure}
\usepackage{cite}
\usepackage{amsmath}
\usepackage{amssymb}
\usepackage{booktabs}
\graphicspath{{fig/}}
\begin{document}

\title{A Tree-structure Convolutional Neural Network for Temporal Features Exaction on Sensor-based Multi-resident Activity Recognition.
\thanks{}}
%
%
\author{Jingjing Cao, Fukang Guo, Xin Lai, Qiang Zhou and Jinshan Dai}
%
%
\institute{School of Logistics Engineering, Wuhan University of Technology, Wuhan, China}
\maketitle              
\begin{abstract}

With the propagation of sensor devices applied in smart home, activity recognition has ignited huge interest and most existing works assume that there is only one habitant. While in reality, there are generally multiple residents at home, which brings greater challenge to recognize activities. In addition, many conventional approaches rely on manual time series data segmentation ignoring the inherent characteristics of events and their heuristic hand-crafted feature generation algorithms are difficult to exploit distinctive features to accurately classify different activities. To address these issues, we propose an end-to-end Tree-Structure Convolutional neural network based framework for Multi-Resident Activity Recognition (TSC-MRAR). First, we treat each sample as an event and obtain the current event embedding through the previous sensor readings in the sliding window without splitting the time series data. Then, in order to automatically generate the temporal features, a tree-structure network  is designed to derive the temporal dependence of nearby readings. The extracted features are fed into the fully connected layer, which can jointly learn the resident labels and the activity labels simultaneously. Finally, experiments on CASAS datasets demonstrate the high performance in multi-resident activity recognition of our model compared to state-of-the-arts techniques.

\keywords{Multi-resident activity recognition \and  CNN \and Feature generation.}
\end{abstract}

\section{Introduction}

Human Activity recognition (HAR) at smart home has emerged as a popular topic to ubiquitous computing with the vast proliferation of sensor devices and Internet of Things \cite{1}. Ambient sensor is the most commonly used with more privacy preservation than camera and less intrusive than wearable sensor. In practical scenarios, there are more than one resident at home, and ambient sensors can only provide binary information with ON or OFF which poses a great challenge to recognize their activities.

In traditional approaches, there are three main steps in HAR: data pre-processing, human activity feature generation and activity classification. The time series data will be first segmented into slices and some statistical features will be extracted. They view each slice as an event ignoring the inherent continuity of events. Advanced heuristic hand-crafted features are generated based on processed features. Inputting those advanced features vectors, a classifier will be trained through machine learning. However, artificially cutting the data into sequences is easy to lead to the disappearance of advanced information. And methods such as Markov models \cite{2} and support vector machines (SVMs) \cite{3} for activity recognition basic signal statistics and waveform traits such as mean and variance of time-series signals to manually generate features is limited by human domain knowledge\cite{4}. There is currently no systematic feature extraction framework to effectively capture distinctive features of human activity. 

Recent years have witnessed the rapid development of deep learning, achieving unparalleled performance in HAR for their favorable advantages of automatically feature learning \cite{5}. The end-to-end learning framework facilitates the training procedure and mutually promotes the feature learning and recognition processes. Temporal convolutional neural networks (TCNNs) recently have been used for time series-processing in sensor-based HAR\cite{6,7}. TCNNs segment the time series data into subsequences, and a variety of processing units can yield an effective representation of local salience of the signals. The deep network allows multiple layers of these processing units to be stacked, so that this deep learning model can characterize the salience in different scales. However, sharing parameters across time is insufficient for capturing all of the correlations between input samples. Additionally, local connectivity limits the output to a function of a small number of neighboring input samples. Another deep neural network recurrent neural networks (RNNs) learn the sequence transfer pattern of the sensor signal well, thus they are widely used to capture temporal dependency for HAR \cite{8,9}. While RNNs only extracts information from the current input, and there is only one shared parameter matrix and difficult to extract deeper representation than TCNN. In addition, little has been learnt in multi-resident field and the existing work rarely make full use of temporal information. Hence, in this work, we proposed a Tree-Structure CNN framework for Multi-Resident Activity Recognition (TSC-MRAR). This approach treat each sample as an event to do predict task, and it can automatically captures neighboring temporal features through previous events and generates deep temporal representation in different dimensions with parameters adaptedly learning in each sliding window.  The contributions of this work are as follows:

\begin{itemize}
\item A  novel end-to-end framework TSC-MRAR for multi-resident activity recognition from binary ambient sensor is proposed, which can automatically generate temporal features leveraging previous events and enable us to classify both residents and activities' labels simultaneously.

\item We  put forward a tree-structure network  for  exploiting deep discriminative features with temporal dependence. Each two features is stacked together and then prorogated to the next layer, which can transmit temporal information to the top embedding, so that we can finally obtain an deep feature representation from neighboring events.

\item Experimental results utilizing multi-resident dataset demonstrate that our approach significantly outperforms state-of-the-arts techniques in multi-resident activity recognition tasks.

\end{itemize}
%
%
%
%

%

The rest of the paper is as follows: In Section II, we review the related work of human activity recognition. In Section III, we elaborate the details of our proposed approach, including deep temporal feature extraction and fully connected layer for classification recognition. We thoroughly evaluate the performance of the proposed framework in Section IV. Finally, we conclude the paper and the future directions in Section V.
\section{Related Work}


\subsection{Deep Learning for HAR}

Deep learning has significant impact on HAR, which benefits for automatically feature extraction procedures for whose lack of the robust physiological basis\cite{18}. In most of existing work on CNNs for human activity recognition, 1D convolution operation is applied to individual univariate time series to capture local dependency along the temporal dimension of sensor signals.  In reference \cite{19}, CNN is applied to human activity recognition using sensor signals. However, this work only made use of a one-layered CNN architecture, which in turn did not exploit the hierarchical physiognomy of activities. In \cite{20}, a three CNN layer is implemented to exploit temporal features. Yang et al.\cite{6} adopts CNN to automate feature learning from the raw inputs in a systematic way. They redefined CNN operators and pooling filters. CNN operators with the same parameters were applied on local signals at different time segments for temporal information. The huge number of possible CNN designs resulting in wasting time and end up building suboptimal solutions after significant amounts of trial-and-error, Baldominos et al.\cite{7} proposed a method to automatically infer the topology of the CNN by an evolutionary algorithm. In\cite{22}, 2D convolution and pooling operations are applied to multivariate time series, in order to capture local dependency along both temporal and spatial domains. However, the size of convolutional kernels restricts the captured range of dependencies between data samples. Typical models are unadaptable to a wide range of activity recognition configurations and require fixed-length input windows.

RNNs perform well in capturing time dependence in lots of researches. In \cite{23}, two representative features are extracted from different statistical profiles and a RNN model is utilized to recognize different human activities. To exploit internal memories of activity sequences, Tran et al.\cite{24} employ multi-label RNN for modelling activities of multiple residents, and recognize activities for each resident separately. Pawlyta et al.\cite{8} introduce models that are able to classify variable-length windows of human activities which is accomplished by utilizing RNNs capacity to read variable-length sequences of input samples and merge the prediction for each sample into a single prediction for the entire window segment. Francisco et al.\cite{9} proposed a  framework composed of CNN and LSTM layers, that is capable of automatically learning feature representations and modelling the temporal dependencies between their activation. However, RNN is limited in its structure that make it difficult to exploit deeper representation than CNNs. 
\subsection{Multi-resident Activity Recognition}

Compare to single-resident activity recognition, multi-resident activity recognition is more realistic and complex. Solutions proposed to solve the problem of sensor-based multi-resident activity recognition can be categorized as knowledge-driven and data-driven models.

Knowledge-driven models are proposed and proven to be effective in single resident activity recognition\cite{10}. Ye and Stevenson \cite{11} proposed a knowledge-driven model named (KCAR) to recognize multi-resident concurrent activities successfully. They integrate a knowledge base and the statistical techniques to segment a continuous sensor sequence into partitions, then transform the problem into single resident activity recognition. Alam et al. \cite{12} present a model to improve the recognition of complex daily activities. They develop a loosely-coupled hierarchical dynamic Bayesian network to identify coarse-grained activities using fine-grained atomic actions and sensor data. In \cite{13}, Jianguo Hao et al. Propose a solution-based on formal concept analysis (FCA) to identify human activities and extract the ontological correlations among sequential behavioral patterns.

Data-driven models rely on data to construct the activity model\cite{14}, and some statistical and probabilistic theories are applied, such as hidden Markov models (HMMs)\cite{2}, conditional random fields (CRFs)\cite{15}. Based on HMM method, many variants of it are proposed. In \cite{16}, Chiang et al. adopted two graphical models, parallel HMM and coupled HMM. Besides, they proposed a new dynamic Bayesian network which extends coupled HMM by adding some vertices to model both individual and cooperative activities. Benmansour et al. \cite{17} developed an HMM-based method, where activities of all residents are linked at each time step to deal with parallel activities and cooperative activities.

However, little has been learnt in multi-resident field to the best of our knowledge in deep learning for HAR. Hence, in this work, we propose an end-to-end framework to automatically capture features for multi-resident activity classification.

\section{TSC-MRAR Model}

\subsection{Problem Statement}

HAR aims to understand human behaviors which enable the computing systems to proactively assist users based on their requirement. Let $S$  be the set of $N$ sensors $S=\left\{S_{1}, S_{2}, \ldots, S_{N}\right\}$. There is a sequence of sensor reading that captures the activity information $x=\left(x_{1}, x_{2}, \ldots, x_{t}\right)$ , where $x_{t}$  denotes the sensor reading at time $t$, which is an event in this work and $x_{t} \in R^{N}$.  $Y_{t}=\left(y_{1}^{t}, y_{2}^{t}, \ldots y_{i}^{t}, \ldots, y_{m}^{t}, y_{m+1}^{t}, \ldots y_{j}^{t}, \ldots, y_{m+n}^{t}\right)$ is the label referring to $x_{t}$, where $m$, $n$ are the number of residents and activities, respectively. $y_{i}^{t}$ is 1 or 0, 1 denotes the certain resident , otherwise, 0 means not. Similarly, $y_{j}^{t}$  is 0 or 1, and 1 denoted the certain activity. Hence the HAR problem can be defined as: given a sequence of sensor reading $x$, we train a network to jointly predict resident and activity label $Y_{t}$ for the event $x_{t}$.

\subsection{Framework Overview}

\begin{figure}[htbp]
\centerline{\includegraphics[width=1\columnwidth]{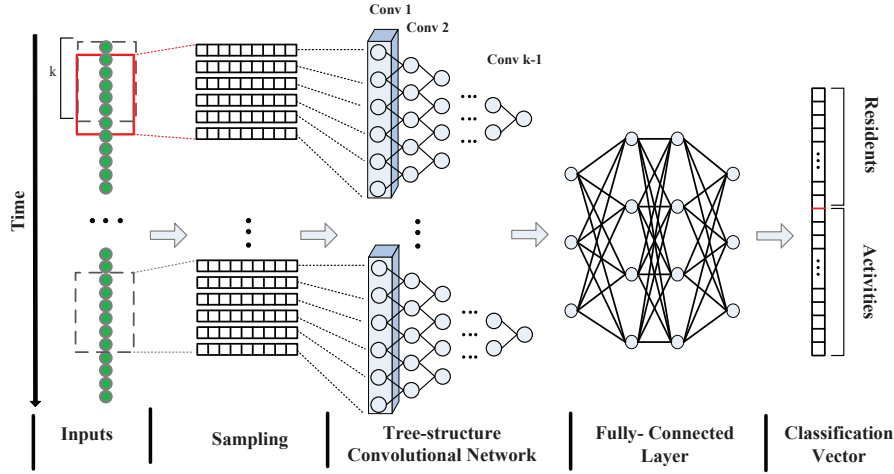}}
\caption{Algorithm flow chart}
\label{fig1}
\end{figure}

In this section, we elaborate the framework of the proposed approach.  TSC-MRAR mainly contains three components: sampling, tree-structure convolutional layer and fully-connected layer for classification recognition. The overall framework is shown in Fig.\ref{fig1}. The input time series data is sampled in a fix sliding window. Those events are embedded into initial vectors, and then are fed into a tree-structure convolutional network, which will output high-order information representation. Each convolutional layer is constructed into a residual block. Finally, top fully connected layer finally combine there embedding  to classify of the labels of residents and activities in the meanwhile.

\subsection{Tree-structure Convolutional Network}

\subsubsection{Sampling}

Traditional deep learning approaches normally segment series data into several slice, and view each slice as an event, while in this work, we treat each ON as an event to do predict task. For each task, we conduct a sample window with fixed size to get previous informations of event in time series so that we can extract the sample feature of this task.  Sampling set is denoted as $X_{t}=\left(x_{t-k}^{0}, \ldots, x_{t-1}^{0}, x_{t}^{0}\right)$, specifically, $k$ is fixed sliding window size. To further describe the proposed model, we define the original input vector as $x_{t}^{0}$ and output for each layer of this network denote as  $x_{t}^{i}$. $X_{t}$ is fed to our network, and output an feature embedding considering to event $x_{t}$.

\subsubsection{Convolutional Operators}
In this section, we elaborate convolutional operators in each layer. There are $k-1$ layers of convolution operations in total. In $i$th layer, the convolution operator is denoted as Conv~i. Set input channel of the first layer convolutional operator Conv~1 to be 1, output channel to be 16, which represents 32 feature maps and convolution kernel size to be 3. Input channel of Conv~2 is 16 and output channel is set to be 32.  The third-layer convolution operator Conv~3 leverages the output feature vectors of Conv~2 with output channel to be 64. The latter $k-4$ convolution operators are the same as Conv~3. A feature map $j$  in $i$ th convolutional layer, denoted by  $z_{i, j}$ is computed as:

\begin{equation}
z_{i, j}=\sigma\left(\sum_{m=1}^{M} w_{m} z_{i-1, j+m}+b_{i-1, j}\right)\label{eq1}
\end{equation}

\noindent where $\sigma(\cdot)$ is the rectified linear unit (ReLU),   $M$ is the size of convolution kernel, $w_{m}$   is the weight matrix and $b_{i-1, j}$  is the bias for this feature map.



\subsubsection{Tree-structure CNN Network}

 Adjacent events can provide with sequence temporal information. The tree-structure CNN network is proposed to capture temporal dependence. Events contextual information is prorogated by stacking two features together and pass between different layers in the tree-structure network. 
 $t$ events in the sampling set $X_{t}$ is fed into the first convolutional layer. Each event embedding will first be convoluted in current layer and then stacked into an feature. In this way, temporal information in $X_{t}$ is continuously transmitted to the next hidden layer.  The final output after $k-1$ layer $x_{t}^{k-1}$  is represented with high-order information. The method of stacking two feature together in each layer is called the basic module.

\begin{figure}[htbp]
\centerline{\includegraphics[width=0.8\columnwidth]{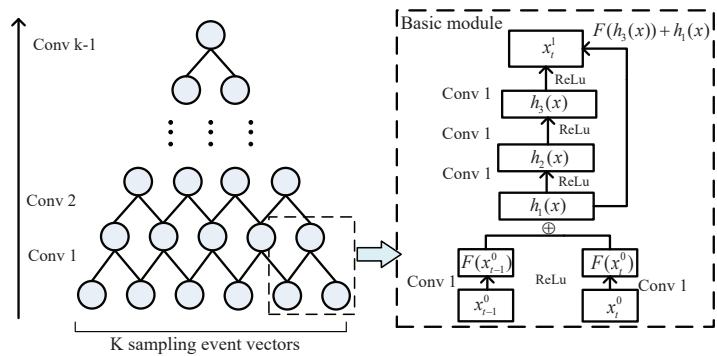}}
\caption{Tree-structure CNN network}
\label{fig3}
\end{figure}

In the basic module,  a residual block is constructed to improve training effect without additional parameters. The Residual block is realized by shortcut connection, and the input and output of this block are overlapped by element-wise through shortcut. As depicted in Fig.\ref{fig3}, it's a 5 layers tree-structure network and symbol $\oplus$ in the figure represents element-level concatenation. The two inputting events embedding $x_{t-1}^{0}$  and  $x_{t}^{0}$ in the sampling set $X_{t}$ is fed into  the basic module.  $x_{t-1}^{0}$  and  $x_{t}^{0}$ is convoluted with Conv~1 and mapped out to be $F\left(x_{t-1}^{0}\right)$ and $F\left(x_{t}^{0}\right)$, respectively. These two feature map are concatenated into a feature vector $h_{1}(x)$ in an element-wise level, and then fed into the residual block. In the residual block, $h_{1}(x)$ will be convoluted for three time with Conv~1 and go through function ReLu to map out as $F\left(h_{3}(x)\right)$. $h_{1}(x)$ is shortcut connected with $F\left(h_{3}(x)\right)$ as the final feature vector $x_{t}^{1}$ in this basic module. 

Every two features in the tree-structure network are concatenated in the way as depicted in the basic module. Hence, temporal information can be transmitted from bottom to top.

\subsection{Fully-connected Layer}

In our end-to-end network, features in the previous convolutional layer will be directly fed into a fully-connected layer. $t$ events in the sampling set $X_{t}$ stack into an feature embedding with high information via tree-structure network. After convolutional layers have extracted relevant features from inputting sampling set, these features can be introduced to a classifier. Taking the extracted feature vectors as the input of CNN feature extracted layer, the fully connected layer trains the data, and classifies different activity types through the learning model.  The final fully connected layer classifies an input batch into one predefined actions and residents classification vector $Y_{t}$. Here we utilize  softmax function as classifier. Once one iteration of forward propagation is done, we will have the error value, with the cross entropy loss function. We are able to use optimizer Adam to update each edge $w$ for the fully connected layer.

\section{Experiments}

\subsection{Dataset}

\begin{figure}[!h]
\centerline{\includegraphics[width=0.8\columnwidth]{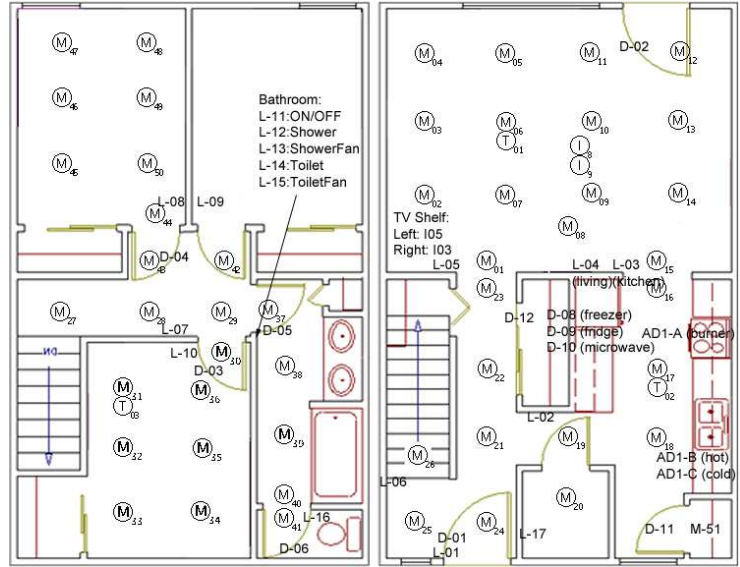}}
\caption{Sensors distribution in the CASAS smart home}
\label{fig4}
\end{figure}

We use a public dataset collected in the CASAS project in the WSU smart workplace in this experiment\cite{25}. There are 15 activities in this activity recognition task with 2 residents, and activities are as shown in Table~\ref{tab1}.  The testbed is equipped with motion and temperature sensors. The motion sensors are located in the ceiling approximately 1m apart and are focused to provide 1m location resolution for the resident. And it uses contact switch sensors to monitor the usage of the phone book, the cooking pot, and the medicine container. There are 37 different binary sensors in total, as depicted in Fig.\ref{fig4}. The events are described by (Date, Time, SensorID, Value, ResidentID, ActivityID). The collected sensor events were manually labeled with the ActivityID and the ResidentID.

\begin{table}[htbp]
\caption{Activities in the CASAS dataset.}\label{tab1}
\centering
\begin{tabular}{|c|c||c|c|}
\hline
\bfseries ID  & \bfseries Description& \bfseries ID  & \bfseries Description\\
\hline
1	 &Filling medication dispenser	 &9	 &Setting table\\
2	 &Hanging up clothes	 &10	 &Reading magazine (R1)\\
3	 &Moving furniture	 &11	 &Paying bills\\
4	 &Reading magazine (R2)	 &12	 &Packing picnic food\\
5	 &Watering plants	 &13	 &Retrieving dishes\\
6	 &Sweeping floor	 &14	 &Packing picnic supplies\\
7	 &Playing checkers	 &15	 &Packing and bring supplies\\
8	 &Preparing dinner	&	&   \\

\hline
\end{tabular}
\end{table}

\subsection{Baselines}
We compare our proposed model with the following methods:

k-Nearest Neighbor (KNN) \cite{27} is a  classification technology  for benchmarking the performance;

Decision Tree (DT)  is also a basic classification method ;

CNN \cite{26} learns MF by optimizing the pairwise ranking loss;

RNN  treat all unobserved interactions as negative samples with smaller weights.
\\

\subsection{Experiment Settings}

Specifically, there are 26 files in the dataset we use, The annotation experiment was conducted by WSU team for 26 times and each experiment was recorded in a single file. We divide 70 percent of them into training set and the remaining into testset, i.e, 18 files are randomly selected as the training set with 6197 events, and the remaining 8 are test sets with 2743 events.

We only utilize the events  with sensor values ON, since there is an time interval in turning on and off the sensor, and these intervals will result in noisy for analysis. Sensors are represented by numbers 0-37 as shown in Table~\ref{tab2} and one hot technology is applied to get initial embedding of each event.  Here we set the parameter $k$ in our approach to be 8.

\begin{table}[!h]
\caption{Tags of sensors.}\label{tab2}
\centering
\begin{tabular}{|cc|cc|cc|cc|}
\hline
\bfseries Sensor ID&  \bfseries Tags & \bfseries Sensor ID& \bfseries Tags & \bfseries Sensor ID&\bfseries  Tags &\bfseries Sensor ID&\bfseries Tags\\
\hline
1& M01&	11& M11 &	 21& M21 	&31& D09\\
2& M02&	12& M12 &	22& M22 &	32& D10\\
3& M03&	13& M13 &	23& M23 &	33& D11\\
4& M04&	14& M14 	&24& M24	&34& D12\\
5& M05&	15& M15 	&25& M25	&35& D13\\
6& M06&	16& M16 	&26& M26	&36& D14\\
7& M07&	17& M17 	&27& M51	&37& D15\\
8& M08&	18& M18     &28& I04&	&  \\
9& M09&	19& M19     &29& I06&	&  \\
10& M10& 20& M20 	&30& D07&	&  \\

\hline
\end{tabular}
\end{table}

The aforementioned comparison methods are divided into two categories, which are traditional machine learning methods and deep learning method. For machine methods, we use the default hyper parameters provided in sci-kit learn package, then we train them for ten times and keep the best result. For deep learning method, RNN is able to predict resident and activty at the same time, but CNN can only predict activity. So we use the same hyper parameters as proposed model and feed them with data preprocessed in the same way.

In detail, we utilize Pytorch 1.2.1 running on python 3.7.4 to construct our networks. We trained our model on server with Ryzen 7 3700x processer and Nvida GTX 1080ti GPU, and the operating system is Ubuntu 18.04 LTS with CUDA 10.0 and cuDNN 7.6.4.

Initially, we adopt 15 epochs for hyper parameters tuning to find appropriate hyper parameters combination of networks. Then we exploit 25 epochs and best hyper parameters combination to train our network, in each epoch, we feed all files in train set into our network and find the average loss to be this epoch's loss output.
%



\subsection{Experimental Results}
We first tune hyperparameters  to rearch the best hyperparameters of our model. Ten-fold cross-validation for evaluation is adopted in our experiment, and we compare the performance of our proposed model with baseline and other competitive methods mentioned above.  To evaluate the performance of aforementioned models, we leverage binary classification metrics for resident classification, including accuracy, F1-score and precision, and multi-classification metric accuracy for activity classification.
\subsubsection{Sensitivity of Hyperparameters}
\begin{figure}[htbp]
\centering
\subfigure[Effect of batch size on loss]{
\begin{minipage}[t]{0.45\linewidth}
\centering
\includegraphics[width=5.5cm]{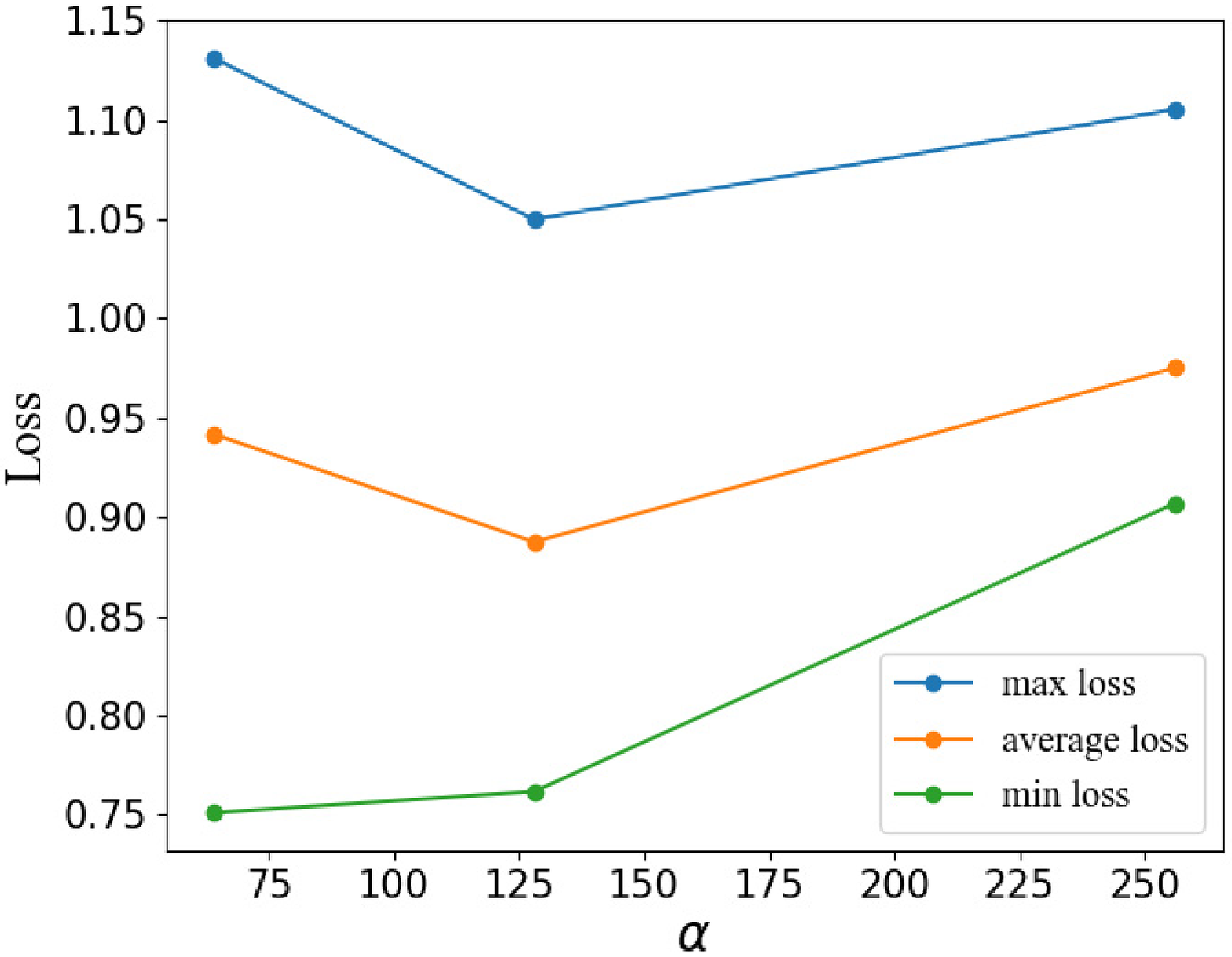}
\label{a}
\end{minipage}%
}%
\subfigure[Effect of L2 and learning rate on loss]{
\begin{minipage}[t]{0.5\linewidth}
\centering
\includegraphics[width=6.5cm]{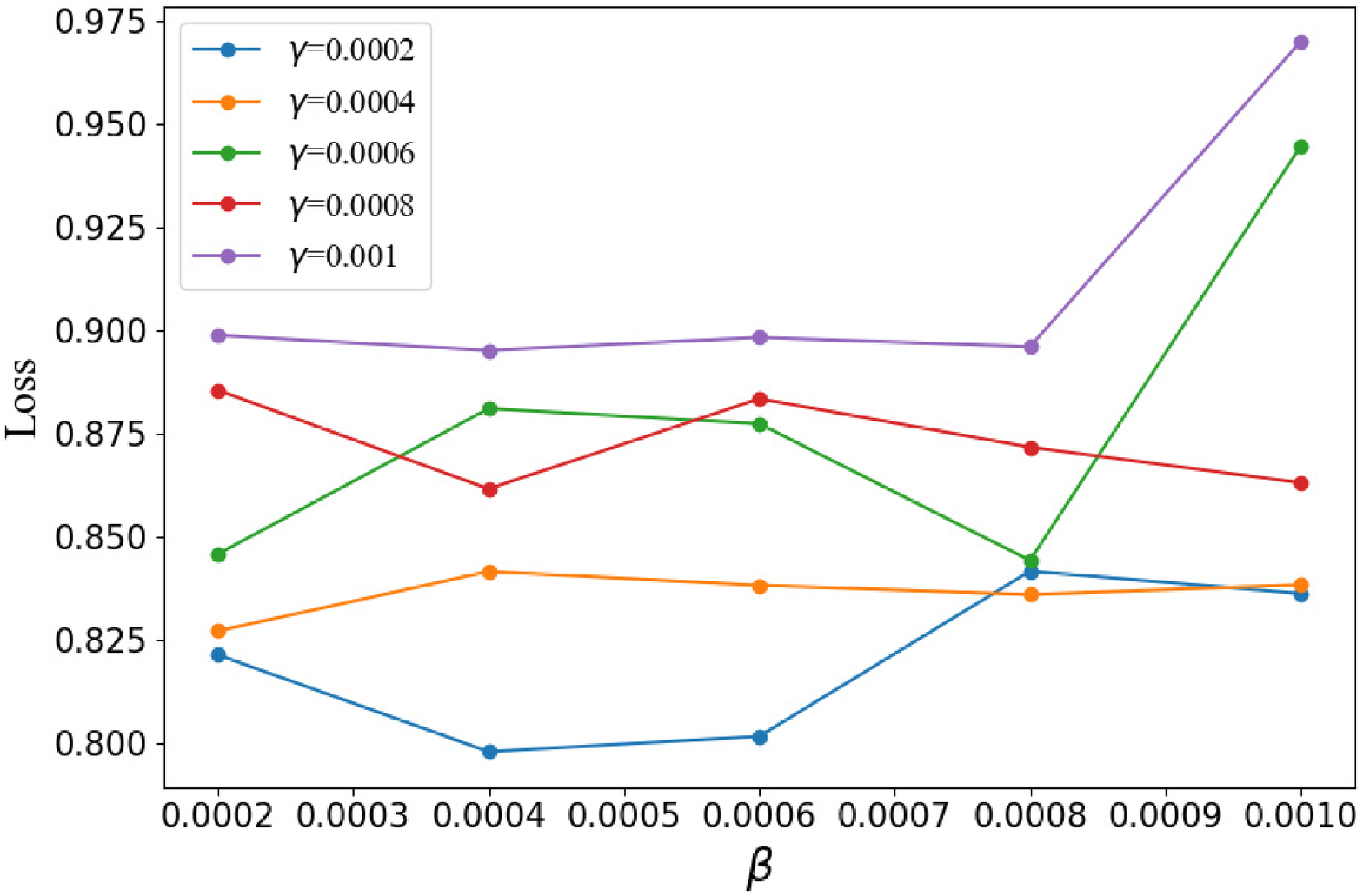}
\label{b}
\end{minipage}%
}%
\caption{Effects of hyper-parameters on loss}
\end{figure}

We characterise the effects of the key hyperparameters of this model, including the batch size $\alpha$, L2 weight $\beta$  and learning rate $\gamma$. The approach of  hyperparameters tuning is to set a few fixed values for the parameters, and then train model to find the optimal parameter combination. Parameter $\alpha$ is preset several discrete values, 64, 128 and 256. $\beta$   and $\gamma$ are both set to range from 0.0001 to 0.001 with step 0.0002. Metrics of max loss, average loss and min loss is utilized to measure batch size performance. From Fig.\ref{a} we can see that $\alpha = 128 $  performs best on its average loss and max loss, while this point is a little bit worse than 64 on min loss. In summary, we pick $\alpha = 128 $. Then we fix $\alpha$ to tune $\beta$  and $\gamma$. Fig.\ref{b} shows that $\beta$  performance in each learning rate setting. When $\gamma$ is set to be 0.0002, $\beta$  can achieve best performance in most case except  $\beta = 0.0008 $ on the overall trend. $\beta$ with value 0.0004 outperform than other values when $\gamma = 0.0002 $. In conclusion, we select $\alpha = 128 $, $\beta = 0.0004$ and $\gamma = 0.0002$ as our final hyperparameters.

\begin{table}[htbp]
\caption{Performance results of TSC-MRAR model and RNN.}\label{tab3}
\centering
\begin{tabular}{|c|c|c|c|c|}

\hline
		&\multicolumn{3}{|c|}{Resident}	&	Activity\\
\hline
 Method  &  Accuracy& Precision  &  F1-score &  Accuracy\\
 \hline
 KNN	 &  0.658&	0.6517&	0.6463	&0.4502 \\
 \hline
 DT	 &  0.7073&	0.6993&	0.7	&0.4765 \\
 \hline
 CNN	 &  -&	-&	-	&0.6593 \\
\hline
RNN	 &  0.7044&	0.6974&	0.6931	&0.3802 \\
\hline
TSC-MRAR	&0.7922	&0.9259	&0.8541	&0.5848\\

\hline
\end{tabular}
\end{table}

%
%


\subsubsection{Results Comparison }

We further compare the best convolutional framework with aforementioned methods as shown in Table~\ref{tab3}.  For resident classification, the TSC-MRAR performed slightly better, though not significantly so.  Our model outperforms most of methods in the table. As for F1-score and Precision, there is a very large improvement of our model compared with the rest methods. Given the complexity of the data with multiple activities, both of the algorithms need huge improvement. CNN get the highest accuracy but cannot recognize resident and activity at the same time, which is extremely important in smart home environment.

In conclusion, these results corroborate our approach perform much better in resident recognition and makes the second best accuracy in activity recognition while CNN get the best result with lack of ability to predict resident at the same time.



\section{Conclusion and Future Works}

In this paper, we consider the problem of multi-resident activity recognition. An end-to-end framework TSC-MRAR to automatically capture temporal features is proposed. Each sample is treated as an event without series data segmentation.  This model can automatically extract temporal features in hidden layers and temporal information is propagated though the tree-structure network. In addition, features extraction and classification are unified in one model to classify multi-resident activities. Experimental results demonstrate that the proposed CNN method outperforms other state-of-the-art approaches, and we therefore believe that the proposed model can serve as a competitive tool of  multi-resident activity classification. 

In the future, we will try to utilize other multi-user data to confirm the wide applicability. A deeper framework  for temporal exaction to explore multi-resident will be concerned to improve the accuracy of multi-resident activity recognition.

\bibliography{ref1}
\end{document}